\documentclass[letterpaper, 10 pt, conference]{IEEEconf}
\IEEEoverridecommandlockouts    

\usepackage{graphicx} 
\usepackage{silence}  							
\usepackage[utf8]{inputenc}    
\usepackage[T1]{fontenc}       
\usepackage[style=ieee,doi=false,isbn=false,date=year,backend=biber,maxbibnames=3,maxcitenames=2,mincitenames=1,uniquelist=false,uniquename=false,giveninits=true,url=false]{biblatex}
\usepackage[nolist,nohyperlinks]{acronym}
\usepackage{subcaption}
\usepackage{graphicx}
\usepackage{booktabs}
\usepackage{amsmath, amssymb}
\usepackage[capitalize]{cleveref}
\usepackage{dsfont}
\usepackage{color}
\usepackage{colortbl}
\usepackage{xcolor}
\usepackage{array}
\usepackage{svg}

\acrodef{MPC}[MPC]{Model Predictive Control}
\acrodef{PID}[PID]{Proportional Integral Derivative}
\acrodef{ADS}[ADS]{Autonomous Driving System}
\acrodef{BEV}[BEV]{Bird's Eye View}
\acrodef{MCC}[MCC]{Meta-Cognitive Component}
\acrodef{NN}[NN]{Neural Network}
\acrodef{VLM}[VLM]{Vision Language Model}
\acrodefplural{VLM}[VLMs]{Vision Language Models}
\acrodefplural{NN}[NNs]{Neural Networks}
\acrodefplural{ADS}[ADSs]{Autonomous Driving Systems}
\acrodef{DPT}[DPT]{Dual Process Theory}

\title{Towards Safer Autonomous Driving in an Open World: A Dual-Process Approach}

\author{Simon Janssen, Michiel Braat$^{1}$, Chris van der Ploeg$^{1,3}$, Serge Thill$^{2}$, Jan-Pieter Paardekooper$^{1,2}$
\thanks{$^{1}$Netherlands Organisation for Applied Scientific Research, Integrated Vehicle Safety Group, 5700 AT Helmond, The Netherlands.}
\thanks{$^{2}$Radboud University, Donders Institute for Brain, Cognition and Behaviour, P.O.Box 9010, 6500 GL, Nijmegen, The Netherlands}
\thanks{$^{3}$Eindhoven University of Technology, Dynamics and Control Group, Mechanical Engineering Dept., P.O. Box 513, 5600 MB, Eindhoven, The Netherlands}}

\begin{document}
\maketitle
\begin{abstract}
    Before autonomous driving systems can be deployed on public roads, it is vital that these systems comply with safety standards, traffic rules, and social norms. Although neural networks trained on large amounts of driving data perform well in routine driving tasks, these models often struggle in novel situations that are not well-represented in the data. In this work, we propose a novel framework that combines a neural network for intuitive, learning-based planning in routine driving tasks with model predictive control for reasoning-based planning in unfamiliar situations, inspired by Dual Process Theory. A meta-cognitive component is designed to switch between the two, using a knowledge graph to reason about contextual risk based on explicit perceptual information and relevant traffic rules and social norms. Contextual risk is represented through risk fields, guiding both the switching mechanism in the meta-cognitive component and compliance with safety standards, traffic rules, and social norms in the reasoning-based planner. The effectiveness of our framework is tested in CARLA for variations of a typical out-of-distribution situations involving (emergency) vehicles running a red light. We show that the novel architecture reduces the number of collisions in the scenarios by 89\% and improves compliance with the special right-of-way rules, compared to the NN-only planner.\\
\end{abstract}

\section{Introduction}
To fully automate driving, \acp{ADS} need to be able to navigate complex and dynamic environments while also ensuring safe and collision-free behavior, obeying traffic rules, and taking into account driving requirements such as passenger comfort and route progression. Advances in machine learning have enabled significant progress in addressing these challenges. While \acp{NN} have been used extensively in perception \cite{pravallika_deep_2024}, they are also increasingly being explored for planning \cite{chen_end--end_2024}. \ac{NN}-based planners can potentially learn more complex and human-like driving behaviors that are difficult to capture directly through hand-crafted rules or objective functions \cite{vitelli_safetynet_2022}. Moreover, \acp{NN} enable end-to-end architectures, which unify perception, motion prediction, and planning 
\cite{codevilla_end--end_2018, toromanoff_end--end_2020} ,
making them less prone to error propagation and misalignment issues arising from inconsistencies between different modules. 

Despite the large potential of \acp{NN} for planning, important challenges remain. One of these challenges is how to improve generalization performance and robustness in out-of-distribution situations. \acp{ADS} will inevitably encounter novel situations due to the complexity and unpredictability of the real world. To prevent serious accidents and fatalities in these situations, safe and predictable behavior is essential. Whereas \acp{NN} excel at finding statistical patterns between input and output, they often struggle to learn the deeper semantic meaning behind this input-output mapping. Hence, when faced with novel inputs, these networks may generalize learned patterns based on semantically irrelevant factors, leading to over-confident or incorrect predictions as a result.

To facilitate adaptivity in the open world and improve safety, we propose a hybrid framework that combines the strengths of \acp{NN} and traditional planners. Specifically, an end-to-end network is used to generate complex and human-like trajectories in familiar situations, while an \ac{MPC}-based planner is used to explicitly take into account contextual information encoded in risk fields to comply with safety standards, legal constraints, and human values in out-of-distribution situations. The integration of the two systems is guided by principles from Dual Process Theory \cite{kahneman_thinking_2011}, which similarly suggests that human intelligence arises from an intuitive system (System I) relying on prior experience, and a reasoning-based system (System II) engaging in more elaborate logical reasoning. To decide when to switch between System I and System II, a \ac{MCC} is designed that uses the same risk fields as the \ac{MPC}-based planner to determine the safety and social desirability of the trajectory predicted by the end-to-end network. The main contributions of this work are as follows:

\begin{itemize}
    \item A hybrid architecture is proposed that combines neural networks and symbolic reasoning to improve generalization and robustness in out-of-distribution situations.
    \item A meta-cognitive component capable of evaluating the safety and social desirability of an end-to-end neural network prediction and decide when to switch to a reasoning-based system.
    \item The effectiveness of our approach is demonstrated through simulation experiments in the CARLA simulator. 
\end{itemize}

\section{Related Work}

\subsection{End-to-End Autonomous Driving}
End-to-end autonomous driving methods have recently emerged as a promising alternative to traditional modular pipelines, which separate perception, planning, and control. The two most prominent end-to-end approaches are imitation learning \cite{ jaeger_hidden_2023, codevilla_end--end_2018} and reinforcement learning \cite{chen_interpretable_2022, toromanoff_end--end_2020}. Several methods directly optimize the planning task, omitting intermediate perception and motion prediction steps \cite{zhang_end--end_2021}. For example, CIL \cite{codevilla_end--end_2018} uses a conditional imitation learning architecture to directly map an input image, input measurements, and a high-level control command to a corresponding output action. Other methods improve interpretability and robustness by learning auxiliary tasks such as object detection, semantic segmentation, depth estimation, and future occupancy prediction \cite{prakash_multi-modal_2021, jaeger_hidden_2023}.

\subsection{Safety in Autonomous Driving}
A popular approach to improve the interpretability and safety of \acp{ADS} is to design explicit rules or cost functions that penalize collisions, deviations from the lane center, and violations of basic traffic rules, as well as other factors related to comfort and progress. Rules or cost functions can be integrated into the model architecture itself, such as in modular end-to-end architectures \cite{hu_planning-oriented_2023, jiang_vad_2023}. They can also be applied externally as a separate safety system \cite{hanselaar_safety_2024}. For example, SafetyNet employs a simple rule-based fallback layer that can modify the predicted trajectory of an end-to-end network when it is dynamically infeasible, violates traffic rules, or has a high likelihood of collision \cite{vitelli_safetynet_2022}. FusionAssurance determines the final safe trajectory using an \ac{MPC} with a cost function that aims to follow the trajectory predicted by an end-to-end neural network, achieve the desired speed, improve comfort, and avoid obstacles modeled with risk functions \cite{zhou_enhance_2024}.

Recently, \acp{VLM} have been explored to act as a safety layer \cite{mei_continuously_2024}. Unlike other methods, \acp{VLM} can leverage their advanced world knowledge and reasoning capabilities to provide more context-aware interpretations of situations \cite{gao_survey_2024}, which can subsequently be used to refine the trajectories generated by an end-to-end or modular architecture \cite{tian_drivevlm_2024}. In this work, we also employ a reasoning-based system to act as a safety layer. However, we do not rely on \acp{VLM}, as these may be slow to respond and may suffer from hallucinations \cite{gao_survey_2024}. Instead, we follow the work in \cite{van_der_ploeg_connecting_2023} and use an \ac{MPC}-based planner and a knowledge graph to generate context-aware trajectories. Although this may result in less human-like behavior, this shortcoming is acceptable when used as a safety system, where the main objective is to ensure safe and socially desirable behavior.

\subsection{Trajectory Evaluation}
A vital aspect of the dual system architecture is the decision when to switch between the \ac{NN}-based planner and \ac{MPC}-based planner. This ties into the concept of meta-cognition, as it requires knowing when the \ac{NN}-based planner cannot make a reliable decision \cite{posner_robots_2020}. Previous work has explored approaches that switch control when a trajectory scoring function exceeds a pre-defined threshold value \cite{hanselaar_safety_2024}. A key advantage of using a trajectory scoring function is that individual predictions can be evaluated on actual measures of risk or reward, rather than approximate measures such as novelty or uncertainty \cite{menda_ensembledagger_2019}. Approximate measures often implicitly assume that end-to-end networks always perform reliably or poorly in certain situations, which may not hold in practice. 

In this work, we design an \ac{MCC} that uses risk fields in the trajectory scoring function to decide when to switch to the \ac{MPC}-based planner. Unlike other trajectory scoring functions \cite{jiang_vad_2023, hu_planning-oriented_2023, hanselaar_safety_2024}, contextual information derived from a knowledge graph is used to shape these risk fields. This allows us to detect not only potential collisions, but also applications of more complex traffic rules and social norms. The proposed trajectory scoring function intentionally excludes comfort and route progression measures, as these factors are not directly relevant to determining the safety of a trajectory and in this work not important enough to switch system. 

\section{Method}
\begin{figure*}
    \centering
    \includegraphics[width=0.7\linewidth]{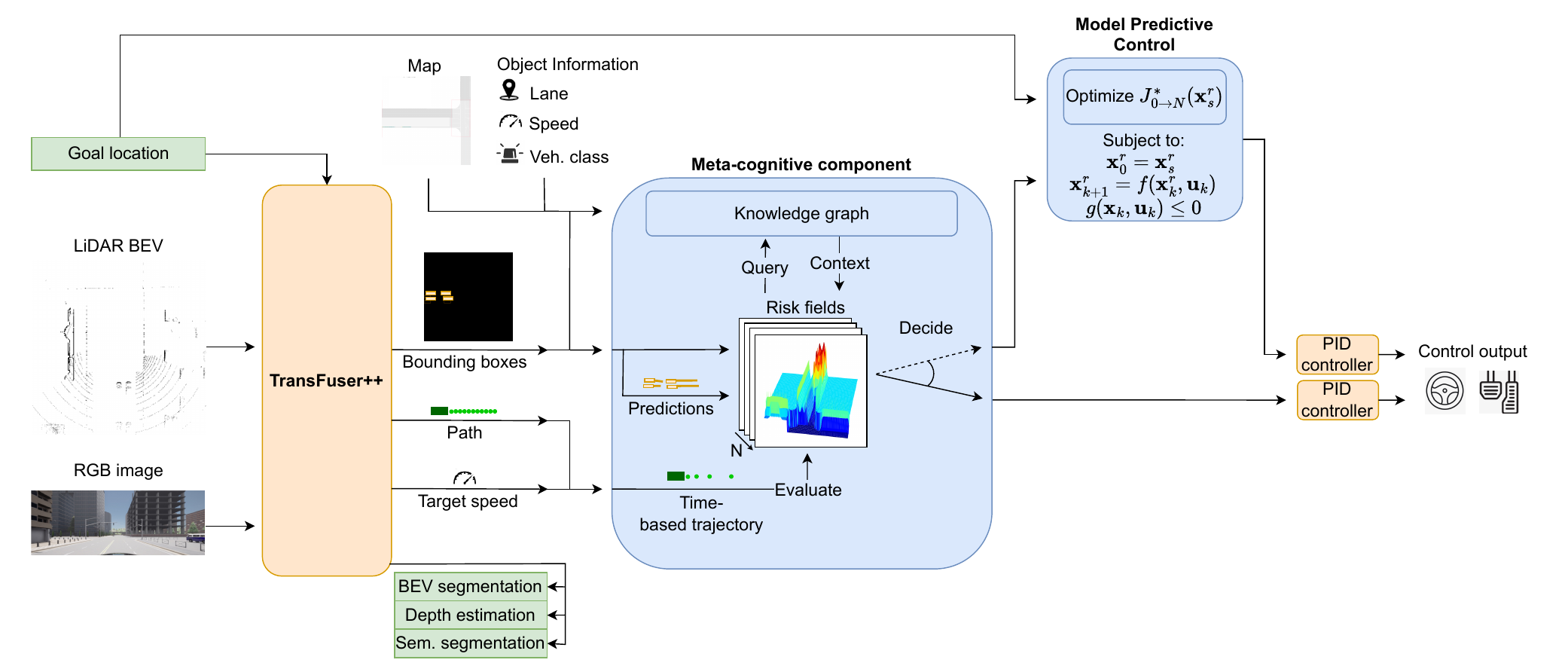}
    \caption{The overall structure of the proposed framework.}
    \label{fig:high-level-architecture}
\end{figure*}

The proposed dual-process framework consists of three main components, as illustrated in Fig.~\ref{fig:high-level-architecture}. A \ac{NN}-based planner is used to predict an initial solution to the planning problem based on available input data from camera, LiDAR, and GPS sensors. This solution is then passed to a \ac{MCC}, which evaluates its safety and desirability and decides whether to initiate a switch to a \ac{MPC}-based planner. When engaged, the \ac{MPC}-based planner can explicitly take into account safety, traffic rules, and social norms encompassed in the risk fields to produce a safer and more socially desirable trajectory at the expense of the ability to generate human-like trajectories.

\subsection{System I (Neural Network)}
In this work, we adopt TransFuser++ as the end-to-end planner, using the codebase and pre-trained model weights made publicly available by the authors \cite{jaeger_hidden_2023}. The model uses a multi-modal fusion transformer to integrate information from a front-facing high-resolution RGB camera (3 x 384 x 1024) and $360^\circ$ field of view LiDAR (voxelized into a 256 x 256 grid). Subsequently, it predicts a target speed and ten future path positions with a fixed distance of one meter between them, conditioned on a goal location (GPS coordinates) provided by a global planner and the current speed. A \ac{PID} controller converts this information to low-level driving commands (steering, throttle, brake). The model learns four auxiliary tasks to improve robustness: semantic segmentation, \ac{BEV} semantic segmentation, depth estimation, and bounding-box object detection. We refer the reader to \cite{jaeger_hidden_2023} and \cite{chitta_transfuser_2023} for more details on the architecture and training procedure. 

\subsection{Meta-Cognitive Component}
The main contribution of this research is the addition of an \ac{MCC} that can determine the safety, legality, and social desirability of the path predicted by the \ac{NN}-based planner, based on contextual information derived from a knowledge graph. 

\subsubsection{Input Description}
Designing a separate perception module that provides perceptual information to the \ac{MCC} is outside the scope of this research. Instead, a combination of stochastic object bounding-box predictions from TransFuser++ and supplementary ground-truth data from perfect sensors in the simulation is used. All perceptual information is expressed in an ego-centric coordinate system, relative to the initial planar state of the ego vehicle given by $(x_0, y_0, \theta_0) = 0$, where $(x_k,y_k)$ represents the center position of the vehicle and $\theta_k$ represents the heading at time step $k$. Objects $\mathbf{o}^{(j)}_k$ can be represented by their kinematic state $(\hat{x}^{(j)}_k, \hat{y}^{(j)}_k, \hat{\theta}^{(j)}_k, v^{(j)}_k, \hat{\delta}^{(j)}_k)$ for all detected objects $j$. $(\hat{x}^{(j)}_k, \hat{y}^{(j)}_k)$ represents the object's predicted position, $\hat{\theta}^{(j)}_k$ represents its predicted heading, $v^{(j)}_k$ represents its ground-truth velocity, and $\hat{\delta}^{(j)}_k$ represents its estimated steering angle (based on the difference between heading and velocity direction). Objects additionally have an associated predicted classification $c^{(j)}$ (where the distinction between regular vehicles and emergency vehicles is based on ground-truth data) and a ground-truth lane assignment $\lambda^{(j)}_k$. Map information is obtained directly from the simulator and covers 32 meters in front and behind the ego vehicle, as well as 32 meters to the left and right. When the ego vehicle moves, the map translates accordingly.

\subsubsection{Knowledge Graph}
To model contextual understanding, relevant information about the current driving situation is organized and linked using a knowledge graph. First, perceptual information is added to the knowledge graph, including object states $\mathbf{o}^{(j)}_0$, object classifications $c^{(j)}$, infrastructure information, and relations between them. Next, reasoning rules are used to add implicit information about traffic rules and social norms. Specifically, rules are used to infer the acceptability of crossing lane markings, the risk levels of objects, the permissibility to drive in a particular lane, and the precedence at intersections. 

\begin{figure}[h]
    \centering
    \includegraphics[width=\linewidth]{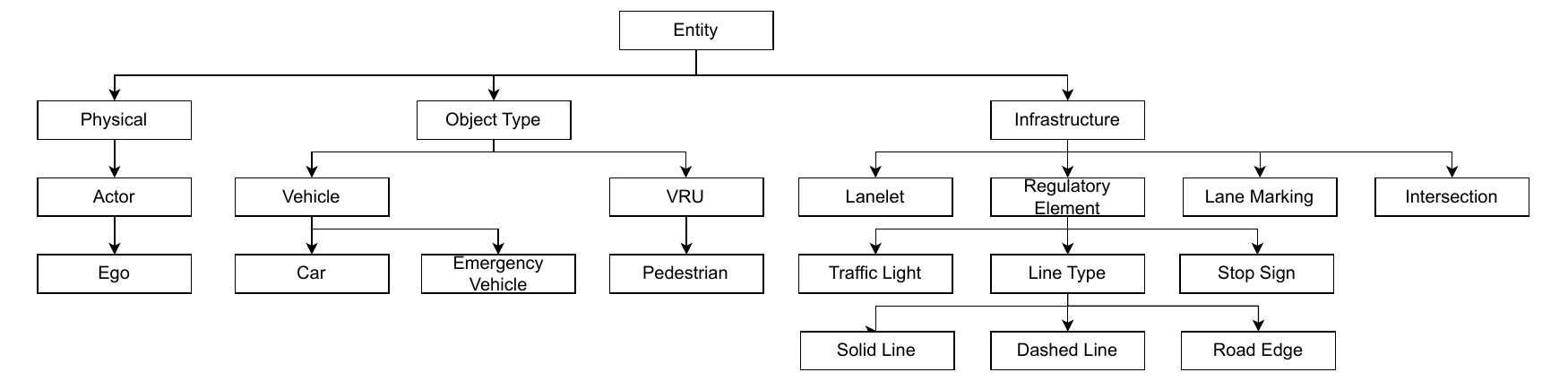}
    \caption{Entity hierarchy.}
    \label{fig:entity-hierarchy}
\end{figure}

\begin{figure}[h]
    \centering
    \includegraphics[width=0.7\linewidth]{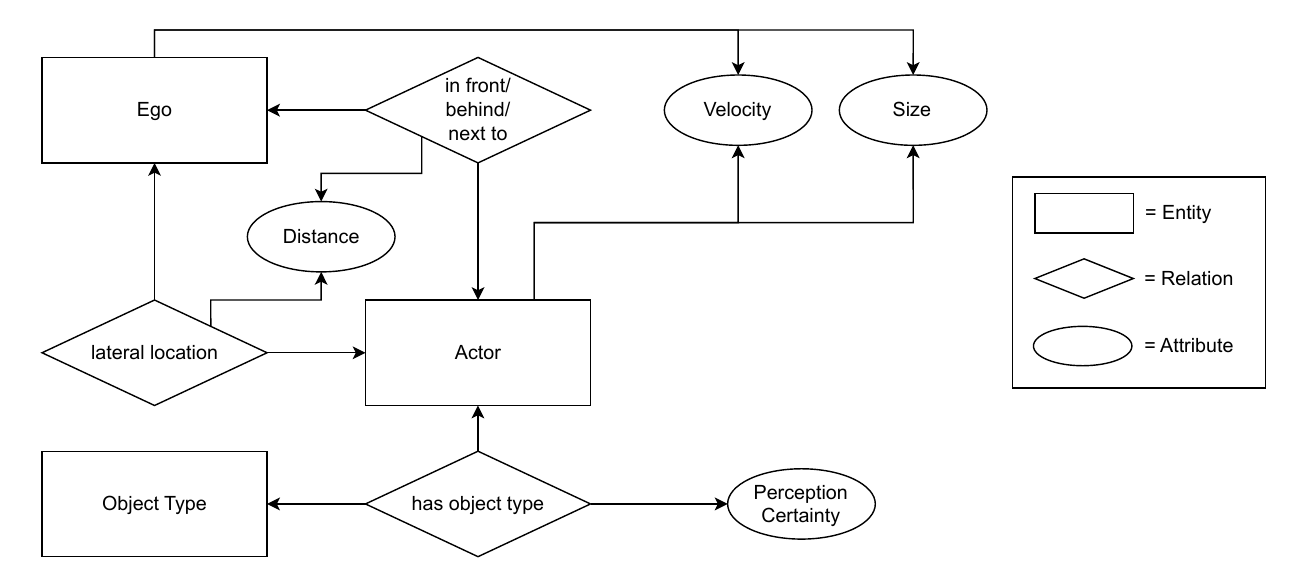}
    \caption{Most important relations for physical entities.}
    \label{fig:entity-relations}
\end{figure}

\begin{figure}[h]
    \centering
    \includegraphics[width=0.7\linewidth]{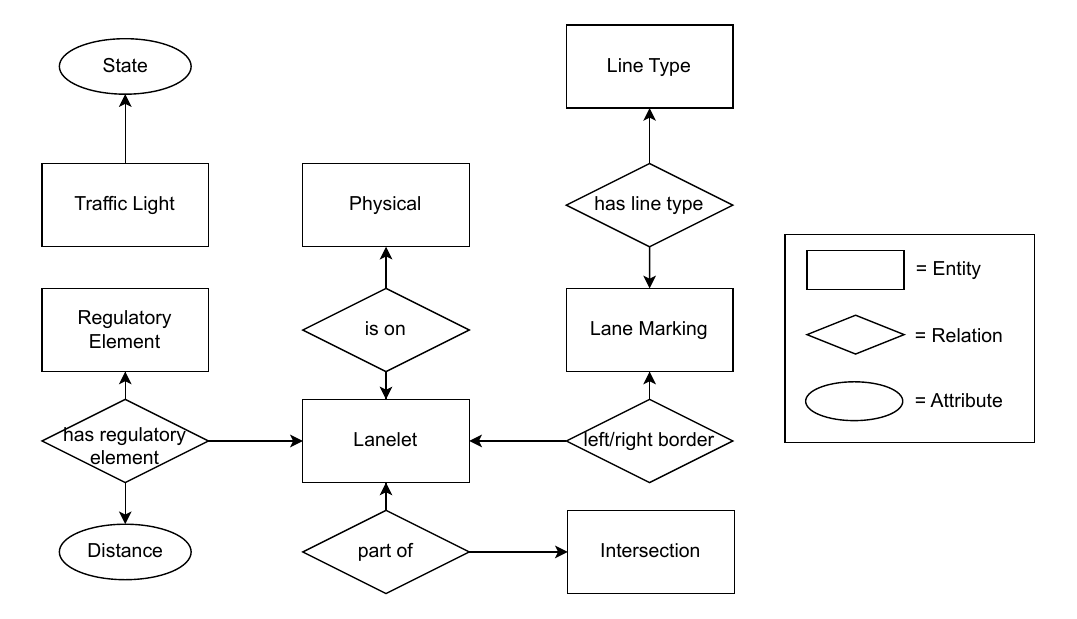}
    \caption{Most important relations for infrastructure entities.}
    \label{fig:infrastructure-relations}
\end{figure}

The knowledge graph and reasoning rules are implemented using TypeDB 3.0. An ontology defines the semantic relationship between concepts through entities, relations, and attributes. Entities represent independent concepts, like detected objects, lanes, or traffic signs. Relations describe interactions between entities by assigning them roles. For instance, the `in front' relation links a subject (ego entity) and an object (actor entity). Attributes assign values to entities or relations, representing properties such as speed or position. Fig.~\ref{fig:entity-hierarchy} shows the complete entity type hierarchy defined in the ontology. Fig.~\ref{fig:entity-relations} and Fig.~\ref{fig:infrastructure-relations} show the most important relation types. 

One of the advantages of using a knowledge graph is that it allows us to derive implicit contextual information based on perceptual information through logical inference. In TypeDB 3.0, this is achieved using functions, which can match patterns in the data to infer additional knowledge.

Following the work of Van der Ploeg et al. (\citeyear{van_der_ploeg_connecting_2023}), functions are used to implement traffic rules and social norms. 
Table~\ref{tab:crossing-acceptability} lists the rules specifying the acceptability to cross lane markings, Table~\ref{tab:object-risk} the rules specifying object risk levels, Table~\ref{tab:driving-acceptability} the rules specifying the permissibility to drive on a particular lane, and Table~\ref{tab:intersection-precedence} the rules specifying the intersection precedence, based on right-of-way rules.
\begin{table}[h]
\centering
\caption{Natural language descriptions and priorities of rules that specify line crossing acceptability.}
\label{tab:crossing-acceptability}
\begin{tabular}{ >{\centering\arraybackslash}p{0.1\linewidth} p{0.8\linewidth} }
\toprule
\textbf{Priority} & \textbf{Natural Language Rule} \\
\midrule
0 & When a line is dashed, it is acceptable to cross \\
0 & When a line is solid, it is not acceptable to cross \\
0 & When a line is a road edge, it is not acceptable to cross \\
1 & When a line is dashed, it is on the left side of the road and connected to a lane in the opposite direction, it is not acceptable to cross \\
2 & When a line is a left or right border of a lane that is itself part of an intersection, it is acceptable to cross \\
3 & When a line is a left or right border of a lane that is itself part of an intersection, and the ego vehicle is on that lane, it is not acceptable to cross \\
\bottomrule
\end{tabular}
\end{table}

\begin{table}[h]
\centering
\caption{Natural language descriptions and priorities of rules that specify object risk levels.}
\label{tab:object-risk}
\begin{tabular}{ >{\centering\arraybackslash}p{0.1\linewidth}  p{0.8\linewidth} }
\toprule
\textbf{Priority} & \textbf{Natural Language Rule}\\
\midrule
0 & An object has medium risk.  \\
1 & When an object is behind the ego vehicle, it has low risk \\
2 & When an object is in front of the ego vehicle,and the object is not on a lane in the opposite direction or a lane that is in the opposite direction of another lane that the ego vehicle can reach by following its current lane, and the ego vehicle could reach the object in less than 2 seconds (assuming constant ego velocity), it has high risk. \\
3 & When an object has an object type emergency vehicle, it has high risk.   \\
\bottomrule
\end{tabular}
\end{table}

\begin{table}[h]
\centering
\caption{Natural language descriptions and priorities of rules that specify driving permissibility.}
\label{tab:driving-acceptability}
\begin{tabular}{ >{\centering\arraybackslash}p{0.1\linewidth}  p{0.8\linewidth} }
\toprule
\textbf{Priority} & \textbf{Natural Language Rule} \\
\midrule
0 & It is permissible to drive on a lane. \\
1 & When a lane is in the opposite direction of the lane that the ego vehicle is currently on, or a lane is in the opposite direction of another lane that the ego vehicle can reach by following its current lane, it is not permissible to drive on.\\
\bottomrule
\end{tabular}
\end{table}

\begin{table}[h]
\centering
\caption{Natural language descriptions and priorities of right of way rules.}
\label{tab:intersection-precedence}
\begin{tabular}{ >{\centering\arraybackslash}p{0.1\linewidth}  p{0.8\linewidth} }
\toprule
\textbf{Priority} & \textbf{Natural Language Rule}   \\
\midrule
0 & The ego vehicle has the right of way at an intersection. \\
1 & When an intersection is managed by traffic lights, the traffic light is turned red, and this is the first intersection that the ego vehicle approaches by following its current lane, the ego vehicle does not have the right of way. \\
1 & When an intersection is not managed by traffic lights, it is the first intersection that the ego vehicle approaches by following its current lane, it is the first intersection that another physical entity approaches by following its current lane, the physical entity comes from the right, and either the physical entity is expected to enter the intersection within 5 seconds, or its distance to the intersection is less than 10 meters, the ego vehicle does not have right of way. \\
2 & When an intersection is the first intersection that the ego vehicle approaches by following its current lane, and it is the first intersection that another physical entity approaches, the physical entity has object type emergency vehicle, and either the physical entity is expected to enter the intersection within 10 seconds, or the distance to the intersection is less than 20 meters, the ego vehicle does not have right of way.  \\
3 & When the ego vehicle is on a lane that is part of an intersection, the ego vehicle has right of way.\\
\bottomrule
\end{tabular}
\end{table}

While the first two rules expand on those provided in \cite{van_der_ploeg_connecting_2023}, the latter two rules are introduced for our specific use case. Since rules may conflict with each other, give rise to exceptions, and have to deal with incomplete knowledge in the open world, they are implemented as default rules with priorities, similar to earlier work \cite{van_der_ploeg_connecting_2023}. At inference time, the rule with the highest priority is selected.

\subsubsection{Populating Risk Fields}
When querying the knowledge graph, the \ac{MCC} obtains implicit contextual information that is used to populate risk potential fields. Two sources of risk are considered in this work: object-imposed risk and infrastructural risk. Similar to \cite{van_der_ploeg_connecting_2023} and \cite{van_der_ploeg_long_2022}, object-imposed risk is described by a bivariate Gaussian distribution:

\begin{table}
\centering
\caption{Risk parameters for objects, based on classification, dimensions, and risk level. }
\label{tab:object-risk-parameters}
\begin{tabular}{l l c c c }
\toprule
\textbf{Object Class} & \textbf{Risk Level} & $A_O$ & $\sigma_x$ & $\sigma_y$  \\
\midrule
Car & Low & 2 & $ 1.5l_O + 0.05$ & $1.1w_O + 0.05$ \\
Car & Medium & 3 & $ 1.5l_O + 0.1$ & $1.1w_O + 0.1$ \\
Car & High & 4 & $ 1.5l_O + 0.3$ & $1.1w_O + 0.3$ \\
Pedestrian & Low & 2 & $ 1.5l_O + 1.2$ & $1.1w_O + 1.2$ \\
Pedestrian & Medium & 4 & $ 1.5l_O + 1.7$ & $1.1w_O + 1.7$ \\
Pedestrian & High & 7 & $ 1.5l_O + 2.2$ & $1.1w_O + 2.2$ \\
Em. vehicle & Low & 2 & $ 1.5l_O + 0.05$ & $1.1w_O + 0.05$ \\
Em. vehicle & Medium & 3 & $ 1.5l_O + 0.1$ & $1.1w_O + 0.1$ \\
Em. vehicle & High & 4 & $ 1.5l_O + 0.3$ & $1.1w_O + 0.3$ \\
\bottomrule
\end{tabular}
\end{table}

\begin{align}
    \label{eq:object-risk}
    &\mathcal{\tilde{U}}^O_k(x,y) = \sum^{N_o}_{n=1} A^{(n)}_O\enspace e^{-\frac{f^{(n)}_o(x, y, \mathbf{o}_k)}{2}}\\
     &f^{(n)}_o = \begin{bmatrix}x - \hat{x}^{(n)}_k & y - \hat{y}^{(n)}_k\end{bmatrix} R \Sigma^{-1}R^{T} \begin{bmatrix}x - \hat{x}^{(n)}_k \\ y - \hat{y}^{(n)}_k\end{bmatrix} \notag \\
     &\Sigma = \begin{bmatrix}(\sigma^{(n)}_x)^2 & 0\\ 0 & (\sigma^{(n)}_y)^2 \end{bmatrix}, \quad R = \begin{bmatrix}\cos \hat{\theta}^{(n)}_k & - \sin \hat{\theta}^{(n)}_k\\ \sin \hat{\theta}^{(n)}_k & \cos \hat{\theta}^{(n)}_k \end{bmatrix} \notag
\end{align}

 where $(x,y)$ represents the evaluation position, $(\hat{x}^{(n)}_k, \hat{y}^{(n)}_k)$ represents the position of the $n$-th object and $\hat{\theta}^{(n)}_k$ represents its heading. $N_o$ represents the total number of objects detected in the environment. The parameters $A^{(n)}_O$, $\sigma^{(n)}_x$, and $\sigma^{(n)}_y$ represent the amplitude and standard deviation of the $n$-th object, which are listed in Table~\ref{tab:object-risk-parameters} and are based on those used in \cite{van_der_ploeg_connecting_2023}.

Infrastructural risk is obtained by either directly mapping the relative positions of infrastructure elements to risk using a binary mask for lanes and off-road risk, or similar to \cite{van_der_ploeg_fear_2024}, by convolving this mask with a 2D Gaussian convolution for lane boundaries and intersections to create a smooth risk map. Each category of infrastructure element $c$ is encoded as a binary mask $M^{(c)}(x, y) \in \{0, 1\}$ over the environmental map, indicating its locations relative to the ego vehicle. Using the binary masks, the risk fields are described as follows:

\begin{table}
\centering
\caption{Risk parameters for lane markings based on the lane type and line crossing acceptability.}
\label{tab:lanemarking-risk-parameters}
\renewcommand{\arraystretch}{1.05} 
\begin{tabular}{l l c c }
\toprule
\textbf{Lane Type} & \textbf{Acceptability} & $A_L$ & $\sigma$  \\
\midrule
Regular & Acceptable & 1.5 & $0.6$ \\
Regular & Not acceptable & 4 & $0.6$ \\
Intersection & Acceptable & 0 & $0.6$ \\
Intersection & Not acceptable & 0.65 & $0.6$ \\
\bottomrule
\end{tabular}
\end{table}

\begin{table}
\centering
\caption{Risk parameters for intersections based on right of way.}
\label{tab:intersection-risk-parameters}
\renewcommand{\arraystretch}{1.05} 
\begin{tabular}{ l c c}
\toprule
\textbf{Precedence} & $A_I$ & $\sigma$ \\
\midrule
Right of way & 0 & 1.5 \\
No right of way & 4 & 1.5 \\
\bottomrule
\end{tabular}
\end{table}

\begin{table}
\centering
\caption{Risk parameters for lanes based on driving permissibility.}
\label{tab:lane-risk-parameters}
\renewcommand{\arraystretch}{1.05} 
\begin{tabular}{ l c }
\toprule
\textbf{Permissibility} & $A_D$  \\
\midrule
Permissible & 0  \\
Not permissible & 3 \\
\bottomrule
\end{tabular}
\end{table}

\begin{align}
    \mathcal{\tilde{U}}^L_k(x,y) &= \sum^{N_l}_{l=1} A^{(l)}_L (M^{(l)}* G^{\sigma^{(l)}})(x, y) \label{eq:lanemarking-risk} \\
    \mathcal{\tilde{U}}^I_k(x,y) &= \sum^{N_i}_{i=1} A^{(i)}_I (M^{(i)}* G^{\sigma^{(i)}})(x, y) \label{eq:intersection-risk} \\
    \mathcal{\tilde{U}}^D_k(x,y) &= \sum^{N_d}_{d=1} A^{(d)}_D M^{(d)}(x, y) \label{eq:lane-risk} \\
    \mathcal{\tilde{U}}^R_k(x,y) &= 4\cdot M^{R}(x, y) \label{eq:offroad-risk} \\
    G^{\sigma^{(c)}}(x, y) &= \frac{1}{2\pi(\sigma^{(c)})^2} e^{-\frac{(x^2 + y^2)}{2(\sigma^{(c)})^2}} \notag
\end{align}

where $\mathcal{\tilde{U}}^L_k$ represents the risk of crossing the boundaries of the lane, $\mathcal{\tilde{U}}^I_k$ represents the risk of driving at intersections, $\mathcal{\tilde{U}}^D_k$ represents the risk of driving in a particular lane and $\mathcal{\tilde{U}}^R_k$ represents the risk of driving off-road. The sign $*$ denotes the 2D convolution with the Gaussian kernel $G^{\sigma^{(c)}}$. The values $N_l$, $N_i$, and $N_d$ represent the number of distinct categories defined for lane markings, intersections, and lanes, respectively. These match the number of entries in Table~\ref{tab:lanemarking-risk-parameters}, Table~\ref{tab:intersection-risk-parameters}, and Table~\ref{tab:lane-risk-parameters}. The terms $A^{(l)}_L$, $\sigma^{(l)}$, $A^{(i)}_I$, $\sigma^{(i)}$, and $A^{(d)}_D$ correspond to the amplitude and standard deviation of the risk fields, which are listed in the same tables for each category and are based on those used in \cite{van_der_ploeg_connecting_2023}. Since lanes and off-road regions are always separated by smoothed lane boundaries or intersections, these infrastructural elements do not require smoothing themselves. Therefore, the convolution with a Gaussian kernel is omitted. 

\subsubsection{Evaluating Predicted Path}
To evaluate risk along the path predicted by TransFuser++, it is converted into a time-based trajectory that encodes the position of the ego vehicle at each time step $k$ for each $k\in[0...N]$, where $N$ is the prediction horizon. Cubic spline interpolation is used to create a smooth curve between the predicted path points. Next, the ego vehicle motion is simulated using a longitudinal PID controller (to predict throttle and brake inputs) and a learned polynomial model (to estimate future velocities). Since the ten predicted future path positions only cover 12.5 meters in front of the ego vehicle, the maximum prediction horizon $N$ dynamically reduces when the ego vehicle is moving fast, such that the final position remains within the predicted 12.5-meter path. Trajectories of surrounding objects are predicted using the Constant Velocity model for pedestrians and the Constant Steering Angle and Velocity model for vehicles \cite{schubert_comparison_2008}.

The risk values of the time-based trajectory prediction of the \ac{NN} model can be obtained by adding the relevant risks from the risk fields:
\begin{align}
    \label{eq:risk}
    \mathrm{risk}_k(x,y) &= \mathcal{\tilde{U}}^O_k(x_k, y_k) + \mathcal{\tilde{U}}^L_k(x_k, y_k) + \mathcal{\tilde{U}}^I_k(x_k, y_k) \notag\\ &\qquad\quad + \mathcal{\tilde{U}}^D_k(x_k,y_k) + \mathcal{\tilde{U}}^R_k(x_k, y_k)
\end{align}

To decide whether to intervene, the maximum risk value along the trajectory is determined (i.e., $\max_{k} \mathrm{risk}_k(x,y)$). When this value exceeds an empirically determined threshold value $T$, the \ac{MCC} initiates a switch to the \ac{MPC}-based planner. 

\subsection{System II (Model Predictive Control)}

The \ac{MPC}-based planner follows the implementation of \cite{van_der_ploeg_long_2022} and \cite{van_der_ploeg_connecting_2023}, optimising a path of minimal risk towards the goal point. The planner uses a non-linear kinematic bicycle model, suitable for urban driving, and input bounds based on the properties of the ego vehicle. The optimization objective of the \ac{MPC} combines this model, the current state, the risk fields (\ref{eq:object-risk}), (\ref{eq:lanemarking-risk}), (\ref{eq:intersection-risk}), (\ref{eq:lane-risk}), and (\ref{eq:offroad-risk}), and a reference state $\mathbf{r}_N$, similar to Eq.~12 in \cite{van_der_ploeg_connecting_2023}. The planar position of $\mathbf{r}_N$ is provided by the global planner (i.e., the same global planner as is used by TransFuser++). Having the risk fields as basis for the optimization objective ensures that the planner takes into account the traffic rules and social norms that form the basis for the risk fields.  

The planner is implemented using the CasADi \cite{andersson_casadi_2019} optimization framework together with the IPOPT non-linear solver \cite{wachter_implementation_2006}. The planner optimizes trajectories over the maximum prediction horizon $N$. For low-level vehicle control, two \ac{PID} controllers are used, identical to the controllers used in \cite{jaeger_hidden_2023}.

\section{Results}
\subsection{Experimental Setup}
To determine the effectiveness of our proposed framework, it is implemented in Python and simulated in CARLA \cite{dosovitskiy_carla_2017}. The simulation parameters are listed in Table~\ref{tab:simulation-parameters}. The risk parameters are listed in Tables~\ref{tab:object-risk-parameters}, \ref{tab:lanemarking-risk-parameters}, \ref{tab:intersection-risk-parameters}, and \ref{tab:lane-risk-parameters}. Risk fields are represented in a 1024x1024 pixel grid map, corresponding to a 64m x 64m area centered on the ego vehicle. For the calculations of the MPC in the CasADi optimizer, the grid map is downscaled by a factor of eight and smoothed using cubic spline interpolation to allow continuous evaluation.  Due to the computational demands of the MCC component and the MPC optimization, the effective simulation timestep is dictated by execution time of the framework, which lead to a rate slower than real time.

\begin{table}
\centering
\caption{Simulation parameters.}
\label{tab:simulation-parameters}
\begin{tabular}{ l l l l}
\toprule
\textbf{Par.} & \textbf{Description} & \textbf{Value} & \textbf{Unit} \\
\midrule
$t_s$ & Sampling time & 0.15 & s\\
$N$ & Maximum trajectory horizon & 20 & - \\
$T$ & Decision threshold MCC & 0.7 & - \\
$\ell_{\mathbf{u}}$ & Input weights & [0.001, 1] & - \\
$m$ & Terminal weights & [0.1, 0.01, 0, 0, 0] & - \\
$\underline{x}, \bar{x}$ & Position x-coordinate & [$-\infty, \infty$] & m\\
$\underline{y}, \bar{y}$ & Position y-coordinate & [$-\infty, \infty$] & m\\
$\underline{\theta}, \bar{\theta}$ & Heading & [$-\pi, \pi$] & rad\\
$\underline{a}, \bar{a}$ & Long. acceleration bounds & [-8, 1.89] & ms$^{-2}$\\
$\underline{\delta}, \bar{\delta}$ & Front wheel angle bounds & [-1, 1] & rad\\
$\underline{v}, \bar{v}$ & Long. velocity bounds & [0, 15] & ms$^{-1}$\\
$\underline{\omega}, \bar{\omega}$ & Steering rate bounds & [-1.5, 1.5] & rads$^{-1}$\\
\bottomrule
\end{tabular}
\end{table}

\begin{figure}[h]
    \centering
    \includegraphics[width=0.75\linewidth]{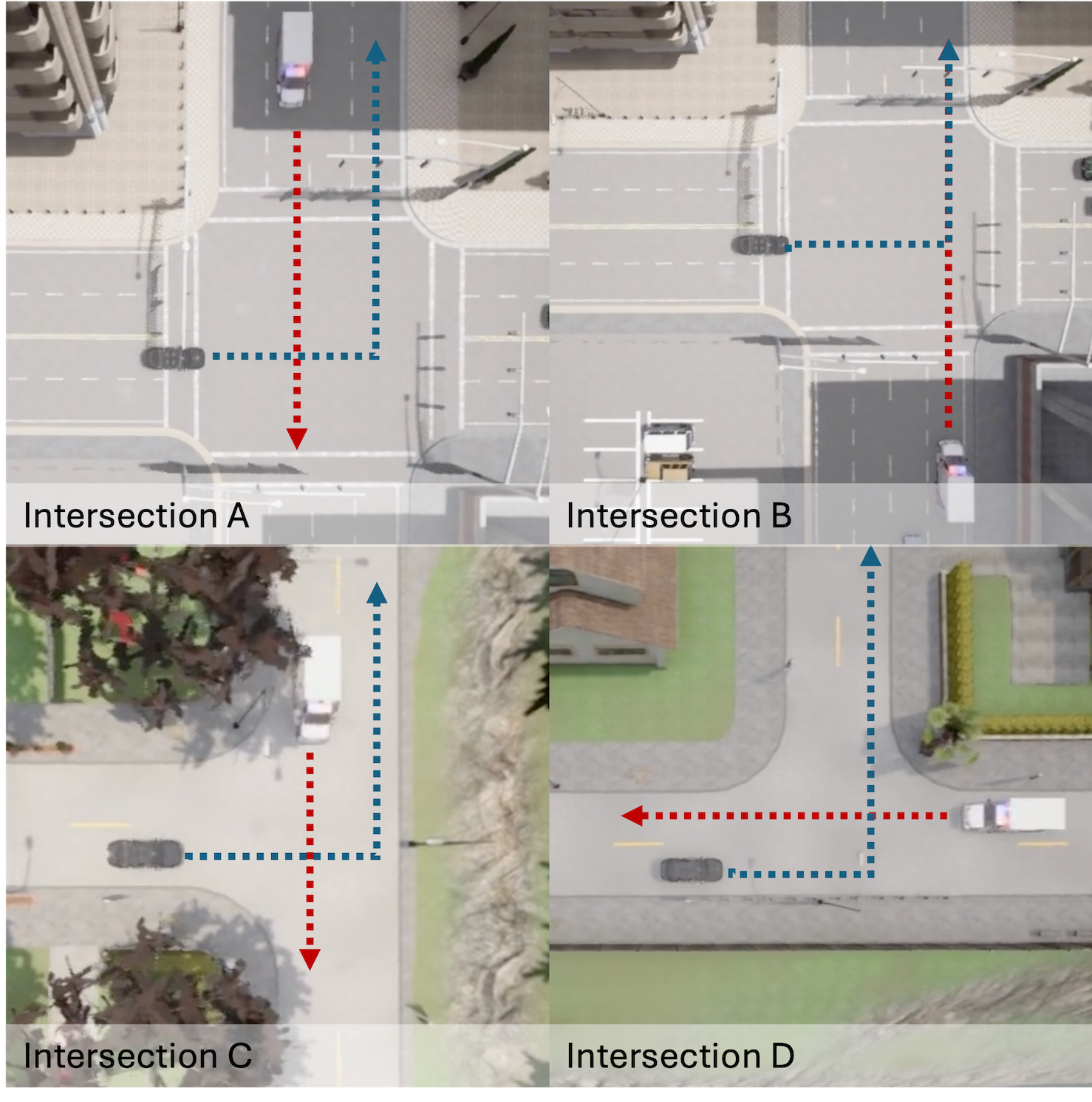}
    \caption{For different intersection setups were used in the simulations.}
    \label{fig:intersections}
\end{figure}

One out-of-distribution scenario category was used in the experiments: an adversary vehicle running a red light at an intersection. This scenario was tested for both a normal car and an emergency vehicle running the red light. In these scenarios, the following parameters were varied: the distance of the ego vehicle to the middle of the intersection (which triggers the start of the scenario), the speed of the adversary vehicle, and the specific intersection at which the scenario occurs. Table~\ref{tab:scenario-parameters} shows the parameters that were used in the scenarios. Fig.~\ref{fig:intersections} shows the different intersections that were used. Scenarios are designed so that the ego vehicle is likely to collide with the adversary vehicle when it does not take any preventive action. Although \acp{NN} are often trained on scenarios where vehicles run a red light, it is a scenario that does not occur frequently, and most existing end-to-end networks 
do not distinguish regular vehicles from emergency vehicles. All parameter combinations were tested, resulting in 160 test scenarios for both the TransFuser++ model with and without the MCC. In addition, the TransFuser v6 (TFv6) resnet34 model \cite{Nguyen2026CVPR} is included in the experiments as a state-of-the-art baseline for comparison. All scenarios were evaluated based on whether the system could avoid a collision with the adversary vehicle on the crossing.

\begin{table}
\centering
\caption{Scenario parameters.}
\label{tab:scenario-parameters}
\begin{tabular}{ l l}
\toprule
\textbf{Parameter} & \textbf{range}\\
\midrule
Scenario trigger distance & [20, 23, 27, 30] (m)\\
Adversary speed &  [35, 40, 45, 50, 55] (km/h)\\
Adversary vehicle model &  [ambulance, car] \\
Intersection & [A, C, B, D]\\
\bottomrule
\end{tabular}
\end{table}

\subsection{Results}
\begin{figure}[h]
    \centering
    \includegraphics[width=1\linewidth]{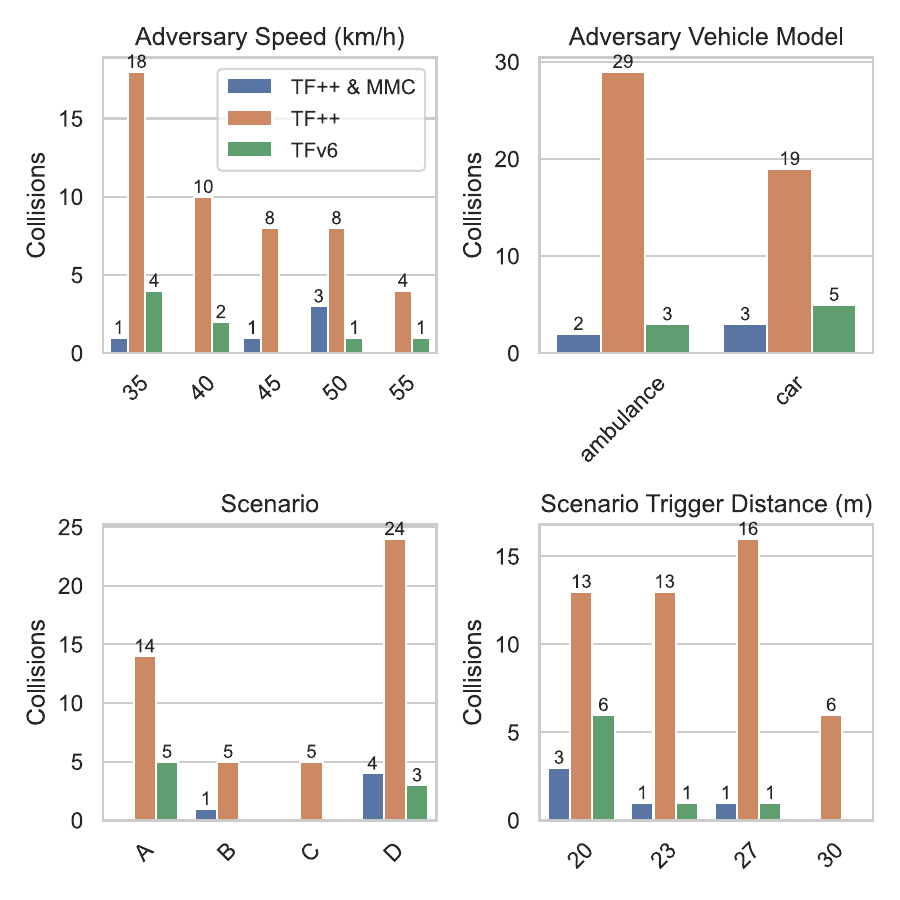}
    \caption{Collisions with vehicle per scenario parameter.}
    \label{fig:collisions}
\end{figure}

\begin{figure}[h]
    \centering
    \includegraphics[width=1\linewidth]{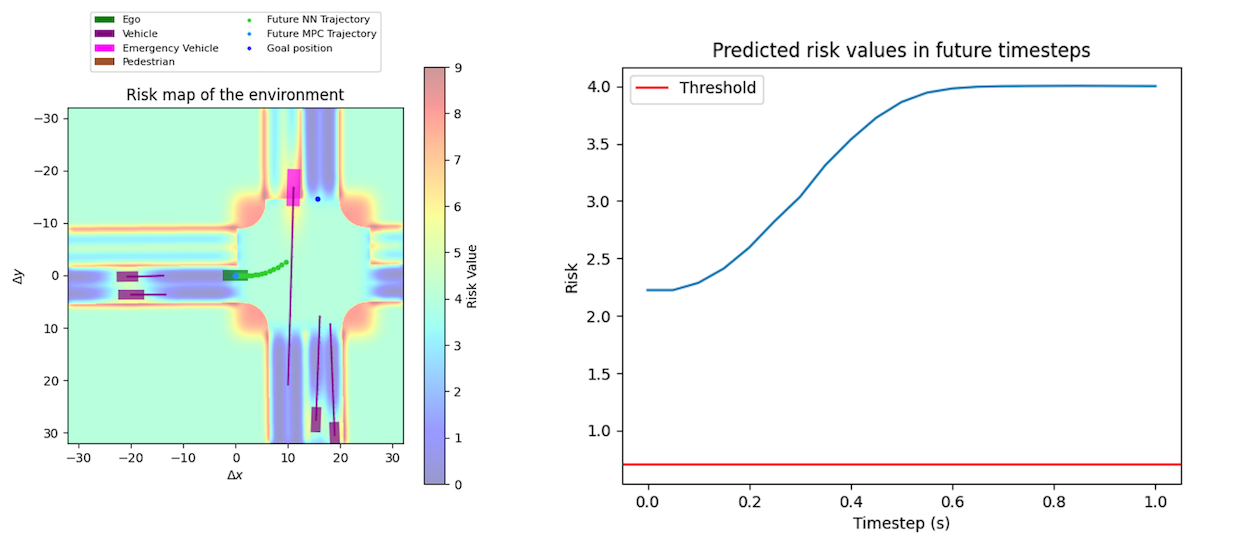}
    \caption{Snapshot where the MCC risk is above the threshold and the MPC is activated, stopping the vehicle.}
    \label{fig:mmc_teshold}
\end{figure}

\begin{figure}[h]
    \centering
    \includegraphics[width=1\linewidth]{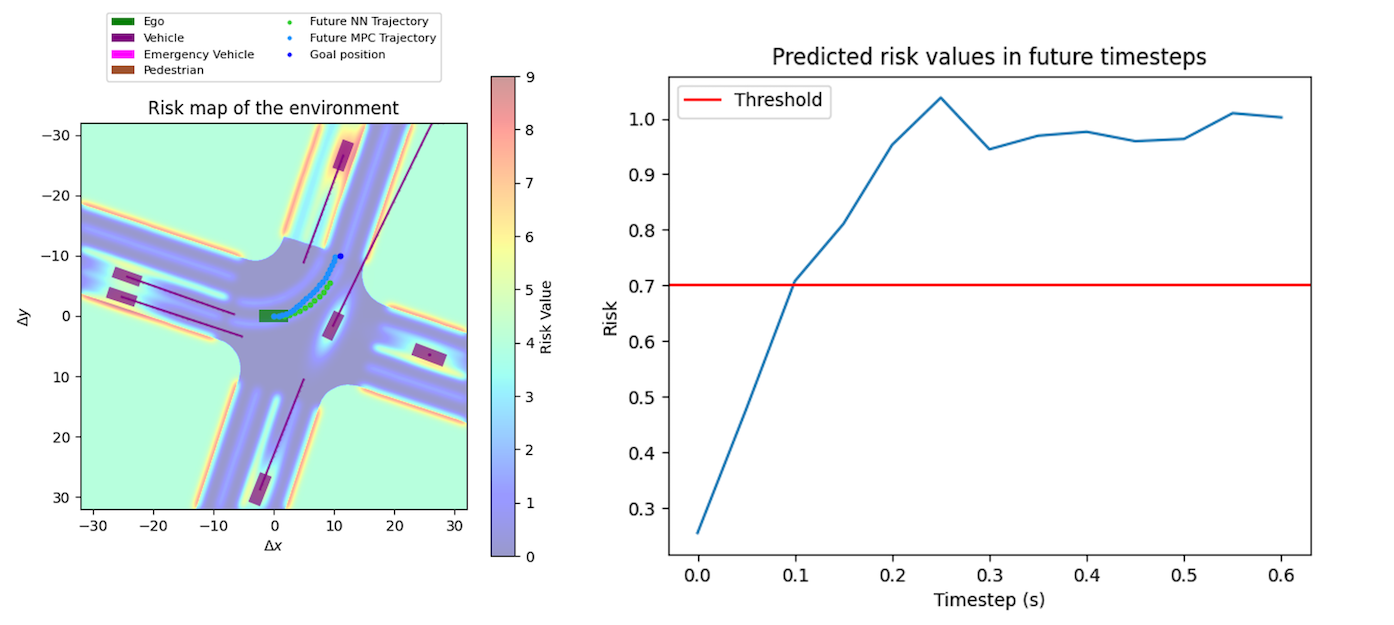}
    \caption{Snapshot where the MCC risk is above the threshold and the MPC is activated adjusting the system I trajectory.}
    \label{fig:mcc_near_collision}
\end{figure}

Collisions that occur in the simulations are shown in Table \ref{tab:result_table}. The TransFuser++ model without the MCC collides with the adversary vehicle in 48 out of 160 simulations, while the system with the MCC results in only 5 collisions. This is similar to the state-of-the-art TFv6 model which had 8 collisions. Fig.~\ref{fig:collisions} shows the number of vehicle-to-vehicle collisions per scenario parameter. As expected, the model with the MCC produces fewer collisions with the adversary vehicle. With the MCC active, the risk of entering the intersection when an ambulance is approaching or when the adversary vehicle is crossing exceeds the threshold, which makes the MPC take over control and stop the vehicle. An example of this is shown in Fig.~\ref{fig:mmc_teshold}. The MCC also enables System II to intervene in near‑collision situations, as the risk around nearby road users becomes sufficiently high. An example of this behavior is shown in Fig.~\ref{fig:mcc_near_collision}. These examples illustrate the benefit of integrating the MCC and reasoning‑based System II: the system explicitly recognizes high‑risk situations or potential violations of traffic rules and, even in out‑of‑distribution cases, prevents unnecessary collisions.

\begin{table}
\centering
\caption{Number of scenarios with infraction.}
\label{tab:result_table}
\begin{tabular}{ l l l l}
\toprule
\textbf{Infraction Type} & \textbf{TF++} & \textbf{TF++ \& MCC} & \textbf{TFv6}\\
\midrule
Collisions with Vehicle & 48 & 5 & 8\\
Collisions with Road Layout &  0 & 7 & 0\\
Outside lane &  0 & 1 & 0 \\
\bottomrule
\end{tabular}
\end{table}

\begin{figure}[h]
    \centering
    \includegraphics[width=0.9\linewidth]{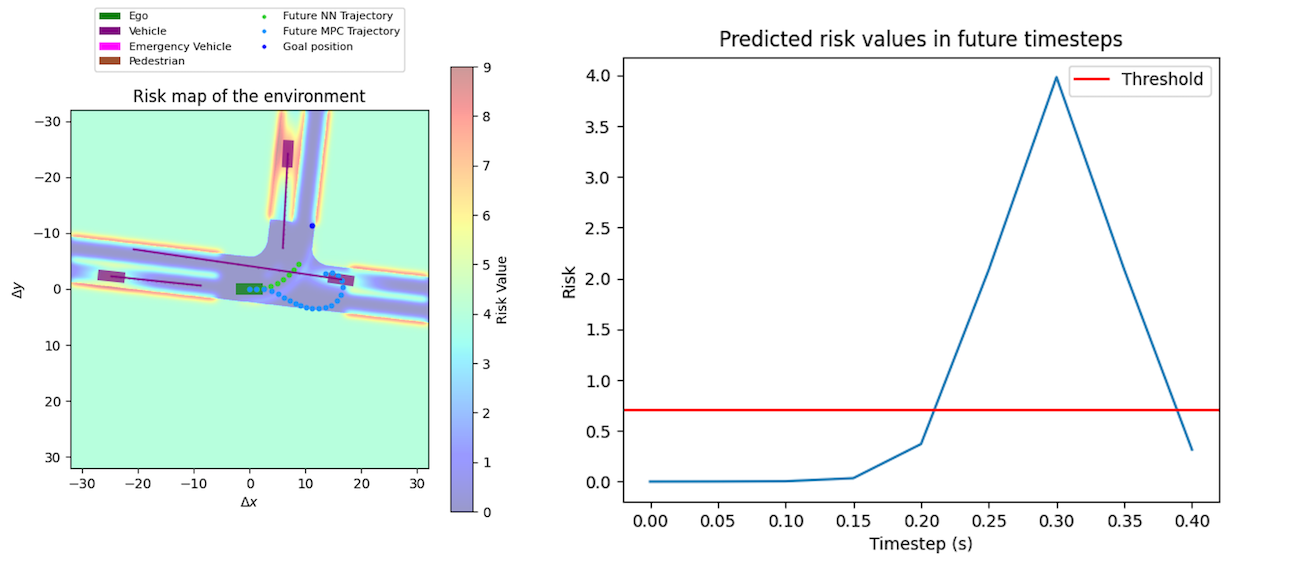}
    \caption{Snapshot where the MPC engages but only is able to calculate a path that collides with a roadside fence.}
    \label{fig:layout_collision}
\end{figure}

Both the TransFuser++ model with and without the MCC perform worst at intersection D. This is likely because, in this configuration, the adversary vehicle approaches from the front rather than from the left or right, making it more difficult for both systems to anticipate and react to the hazard. Furthermore, all road‑layout collisions for the model with the MCC occur in Scenario 4. These collisions happen specifically when the MCC intervenes by steering the vehicle away from the adversary. While this intervention avoids a collision with the adversary vehicle, it causes the MPC-generated trajectory to require a larger rotation angle than the ego vehicle can feasibly follow at that moment. As a result, the vehicle is unable to adhere to these larger rotation angles, drifts outside the lane, and ends up colliding with the tree or fences located on the curb. Figure \ref{fig:layout_collision} shows a timestamp when the MPC intervenes in one of the 7 layout collision simulations. The cause of the mismatch in the intended steering angle by the MPC and the actual angle executed by the PID controller shows the challenge in correctly tuning the MPC.

\section{Conclusions}
In this work, we propose a novel dual-process approach that combines an \ac{NN}-based planner and a context-aware \ac{MPC}-based planner. We designed an \ac{MCC} that uses risk fields shaped by perceptual information, traffic rules, social norms, and predicts trajectories of surrounding objects to evaluate the safety and social desirability of a trajectory generated by an end-to-end neural network. When the risk associated with this trajectory exceeds an empirically determined threshold value, a switch is initiated to an \ac{MPC}-based planner, which explicitly uses the risk fields to generate a safer, although less human-like, alternative trajectory. We tested our method in CARLA through simulations involving an adversary vehicle running a red light at intersections. We show that our method  improves the safety and adaptivity of the \ac{ADS} by reducing the amount of vehicle-to-vehicle collisions in the simulations by 89\%. Future work will be directed towards integrating more advanced motion prediction models, adopting a more principled approach to derive the decision threshold in the \ac{MCC}, improving the computational time and scaling the method to other traffic scenarios. 

\subsection{Acknowledgment}
This work was partly funded by the Horizon Europe programme of the European Union, under grant agreement 101202457 (project AIGGREGATE). Views and opinions expressed here are however those of the author(s) only and do not necessarily reflect those of the European Union or CINEA. Neither the European Union nor the granting authority can be held responsible for them.

\printbibliography

\end{document}